# Learning Hidden Markov Models with Geometrical Constraints

Hagit Shatkay*
National Center for Biotechnology Information, NLM/NIH
Bethesda, MD
shatkay@ncbi.nlm.nih.gov

## Abstract

Hidden Markov models (HMMs) and partially observable Markov decision processes (POMDPs) form a useful tool for modeling dynamical systems. They are particularly useful for representing environments such as road networks and office buildings, which are typical for robot navigation and planning. The work presented here is concerned with *acquiring* such models. We demonstrate how domain-specific information and constraints can be incorporated into the statistical estimation process, greatly improving the learned models in terms of the model quality, the number of iterations required for convergence and robustness to reduction in the amount of available data. We present new initialization heuristics which can be used even when the data suffers from cumulative rotational error, new update rules for the model parameters, as an instance of generalized EM, and a strategy for enforcing complete geometrical consistency in the model. Experimental results demonstrate the effectiveness of our approach for both simulated and real robot data, in traditionally hard-to-learn environments.

## 1 Introduction

Hidden Markov models (HMMs), as well as their generalization to partially observable Markov decision processes (POMDPs), model a variety of nondeterministic dynamical systems as probabilistic state-transition systems with discrete states, observations, and possibly actions. In this paper we concentrate on the special case of models in which states can be associated with points in a metric configuration space. These are appropriate in contexts such as office building, road network, or sewerage system modeling. Specifically, such POMDP models form a useful basis for robot navigation in buildings, providing a sound method for localization and planning [SK95, NPB95, CKK96]. Much of the previous work on planning assumed that the model is acquired manually; such manual acquisition can be very tedious and it is often difficult to obtain correct probabilities.

Learning such models automatically is an ultimate goal, both for robustness and in order to cope with new and changing environments. Since POMDP models are a simple extension of HMMs, they can, theoretically, be learned with a simple extension to the Baum-Welch algorithm [Rab89] for learning HMMs. However, without a strong prior constraint on the structure of the model, the Baum-Welch algorithm does not perform very well: it is slow to converge, requires a great deal of data, and often becomes stuck in local maxima.

Our work focuses on showing how weak information about the metric relationship between states can be used to significantly improve model learning. Such information is usually readily available but is often ignored during the process of learning topological maps. We have previously shown [SK97, SK98] that the odometric ability of the robot, which allows it to roughly measure its geometric position changes while moving in the environment, can be very useful when learning *topological* models.

This paper addresses several issues not previously dealt with: It introduces a "lag-behind" estimation procedure that enforces geometrical constraints, while being an instance of *generalized* EM, new heuristics for choosing an initial model from which the iterative optimization starts, and an update strategy that allows the enforcement of the complete geometrical constraints (*additivity*), while our earlier work enforced only part of the constraints (antisymmetry of the odometry between points). We conclude by empirically demonstrating the effectiveness of our algorithms for learning models of environments that are traditionally considered hard to learn.

## 2 Related Work

The work presented here is concerned with *learning statistical models* in the context of *robot navigation*. In the robotics domain it is common to distinguish between two main types of maps: *geometric* and *topological*. The former represent the environment in terms of the objects placed in it and their positions. For example, *grid-based* maps [ME85, Asa91, TBF98] are an instance of the geometric approach. Such maps are the best choice when it is necessary for a robot to know its location accurately in terms of metric coordinates. However, in our environments of interest such as office buildings with corridors and rooms or networks of roads, *topological* maps [KB91], specifying the important locations and their connections, suffice. Such maps are typically less complex and support much more efficient planning than metric maps.

*This work was supported by the Brown University Graduate Research Fellowship.

We draw an additional distinction, between world-centric[1] *maps* that provide an "objective" description of the environment independent of the agent using the map, and robot-centric *models* which capture the *interaction* of a particular "subjective" agent with the environment. An agent learning a *map* (such as the grid maps mentioned above), takes into account its own noisy sensors and actuators and tries to obtain an objectively correct map that other agents could use as well; other agents need to compensate for their own limitations when assessing their position according to the map. We take the approach of learning a *model* that captures the *interaction* of the agent with the environment. Hence, the noisy sensors and actuators specific to the agent are reflected in the model; this approach allows robust planning, taking into account the error in sensing and action, (although a different model is likely to be needed for different agents). Moreover, topological models support a more general notion of state, possibly including information such as the robot's battery voltage or arm position.

The work most closely related to ours is by Koenig and Simmons [KS96a, KS96b], who learn POMDP models (stochastic topological models) of a robot hallway environment. To overcome the hardship of learning such a model without initial information they use a *human-provided* topological map to start from, and further constraints on the structure of the model. A modified version of the Baum-Welch algorithm learns the parameters of the model. They also developed an incremental version of Baum-Welch that allows it to be used on-line in certain kinds of environments. Their models contain very weak metric information, representing hallways as chains of 1-meter segments and allowing the learning algorithm to select the most probable chain length. This method is effective, but results in large models (size is proportional to hallways' length). In contrast, we directly incorporate odometric information into the Baum-Welch algorithm to learn a probabilistic model with both discrete and continuous probabilities.

Probabilistic models are widely used within the AI community. Such models may allow continuous probabilities, as demonstrated in work on Bayesian networks [HG95], HMMs [GJ97] and stochastic maps [SSC91]. However, that work significantly differs from ours in several ways. Commonly, the continuous distributions used are *linear* – that is – distributions assigning density to each point on the real line so that the area under the density curve, integrated over the whole real line, is 1, (most often the distribution is Gaussian). As pointed out in our earlier work [SK98], directional data is inherently *cyclic*, requiring the use of *circular distributions*, where for some period $\psi$ (a real number), the density of any point $x$ is the same as that of $x+k\psi$, for any integer $k$. In addition, usually the learned statistical parameters are unconstrained (aside for the obvious constraint of *being a distribution*.) Our approach, which enforces geometrical constraints when estimating the parameters, requires special precautions to ensure convergence of the iterative reestimation procedure, as demonstrated in the following sections.

## 3  Models and Assumptions

We describe here the model (and later the algorithms) for learning an HMM, rather than a POMDP. The extension to POMDPs – which we developed and implemented – is technically straightforward but notationally more cumbersome.

The world is composed of a finite set of states, whose number is assumed here to be known. The dynamics of the world are described by state-transition distributions, specifying the probability of transitioning from one state to the next. A finite set of possible observations is associated with each state; the observation frequency is described by a probability distribution and depends only on the current state. In our model, observations are *multi-dimensional*, hence, an observation is a vector of values, each chosen from a finite domain. We assume that observation values are conditionally independent, given the state.

In addition, each state is assumed to be associated with a (not necessarily unique) point in some metric space. Whenever a state transition is made, encoders on the robot's wheels allow it to record its current pose (position and orientation) relative to its pose in the previous state. It is assumed that the position change ($\Delta x$, $\Delta y$) is corrupted with independent 0-mean *normal* noise, while the orientation change, ($\Delta \theta$), is corrupted with independent *von Mises*-distributed noise. The von Mises distribution is a circular version of the normal distribution, and its density function is: $f_{\mu,\kappa}(\theta) = \frac{1}{2\pi I_0(\kappa)} e^{\kappa \cos(\theta - \mu)}$, where $\kappa$ is a concentration parameter and $I_0(\kappa)$ is the modified Bessel function of the first kind and order 0. It is extensively discussed in former work [GGD53, Mar72, SK98].

In early work [SK97] we assumed perpendicularity of the corridors that was taken advantage of while the robot collected the data; Odometric readings were recorded with respect to a *global coordinate system*, and the robot could re-align itself with the origin after each turn. A trajectory of odometry recorded under this assumption by our robot Ramona, along the $x$ and $y$ axes is given in Figure 1. In contrast, Figure 2 shows a trajectory of the odometry recorded *without* the perpendicularity assumption. The data collected under the latter setting is subjected to cumulative rotational error. In recent work [SK98] we have shown how such data can be handled through *state-relative coordinate systems*, as explained later in this section. This solution is reflected both in the constraints imposed on the model and in the learning algorithm.

To state the setting formally, a model is a tuple

---
[1] Thanks to Sebastian Thrun for the terminology.

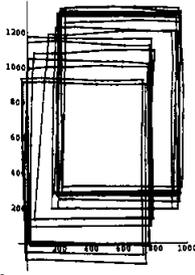

Figure 1: Sequence gathered by Ramona, perpendicularity assumed.

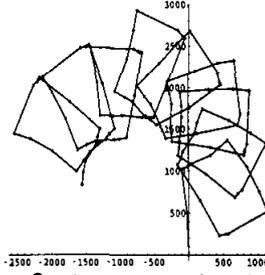

Figure 2: Sequence gathered by Ramona, no perpendicularity assumed.

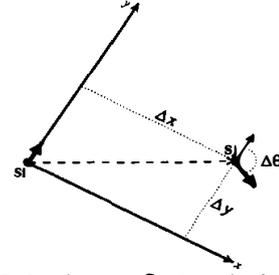

Figure 3: Robot in state $S_i$, faces in the y-axis direction; the relation $S_i, S_j$ is WRT $S_i$'s coordinate system.

$\lambda = \langle S, O, A, B, \pi, R \rangle$, where

- $S = \{s_0, \ldots, s_{N-1}\}$ is a finite set of $N$ states;

- $O = \prod_{i=1}^{l} O_i$ is a finite set of observation vectors of length $l$; the $i$th element of an observation vector is chosen from the finite set $O_i$;

- $A$ is a stochastic transition matrix, with $A_{i,j} = Pr(q_{t+1} = s_j | q_t = s_i)$; $0 \leq i, j \leq N-1$; $q_t$ is the state at time $t$;

- $B$ is an array of $l$ stochastic observation matrices, with $B_{i,j,o} = Pr(V_t[i] = o | q_t = s_j)$; $1 \leq i \leq l$, $0 \leq j \leq N-1$, $o \in O_j$; $V_t$ is the observation vector at time $t$;

- $\pi$ is a stochastic initial probability vector describing the distribution of the initial state; for simplicity it is assumed here to be $\langle 0, \ldots 0, 1, 0, \ldots, 0 \rangle$, implying that the robot always starts in a designated initial state $s_i$;

- $R$ is a relation matrix, specifying for each pair of states, $s_i$ and $s_j$, the *mean* and *variance* of the metric relation between them along the $x$ and the $y$ dimensions; e.g. $\mu_{ij}^x \stackrel{\text{def}}{=} \mu(R_{i,j}[x])$ is the mean of the $x$ component of the relation between $s_i$ and $s_j$, and $(\sigma_{ij}^x)^2 \stackrel{\text{def}}{=} \sigma^2(R_{i,j}[x])$, the variance. As shown in earlier work [SK98] $R$ also contains the *mean* and *concentration* of the *change in heading* between the two states, $\mu_{ij}^\theta$ and $\kappa_{ij}^\theta$. Furthermore, $R$ is *geometrically consistent*: In a *global* coordinate system this means that for each component $w \in \{x, y, \theta\}$, the relation $\mu_{ab}^w \stackrel{\text{def}}{=} \mu(R_{a,b}[w])$ must be a *directed metric*, satisfying the following constraints (referred to as *global constraints* from here on) for all states $a, b$, and $c$:

- ⋄ $\mu_{aa}^w = 0$;
- ⋄ $\mu_{ab}^w = -\mu_{ba}^w$ *(anti-symmetry)*; and
- ⋄ $\mu_{ac}^w = \mu_{ab}^w + \mu_{bc}^w$ *(additivity)*.

In a *state-relative coordinate system* these same constraints apply to the $\theta$ component, but the constraints over $x$ and $y$ need to be specified with respect to the explicit coordinate system used. As shown in Figure 3, each state, $s_i$, has its own coordinate system; the $y$ axis is aligned with the robot's heading in the state (denoted by bold arrows in the figure), and the $x$ axis is perpendicular to it. The geometric relation from $s_i$ to $s_j$ is expressed with respect to the coordinate system of $s_i$.

Given a pair of states $a$ and $b$, we denote by $\mu^{\langle x, y \rangle}(a, b)$ the vector $\langle \mu(R_{a,b}[x]), \mu(R_{a,b}[y]) \rangle$. Let us define $\mathcal{T}_{ab}$ to be the transformation that maps an $\langle x_a, y_a \rangle$ point represented with respect to the coordinate system of state $a$, to the same point represented with respect to the coordinate system of state $b$, $\langle x_b, y_b \rangle$.

More explicitly, let $\mu_{ab}^\theta$ be, as before, the mean change in heading from state $a$ to state $b$. Applying $\mathcal{T}_{ab}$ to a vector $\langle \begin{smallmatrix} x_a \\ y_a \end{smallmatrix} \rangle$ results in the vector $\langle \begin{smallmatrix} x_b \\ y_b \end{smallmatrix} \rangle$ as follows:

$$\left\langle \begin{matrix} x_b \\ y_b \end{matrix} \right\rangle = \mathcal{T}_{ab} \left\langle \begin{matrix} x_a \\ y_a \end{matrix} \right\rangle = \left\langle \begin{matrix} x_a \cos(\mu_{ab}^\theta) - y_a \sin(\mu_{ab}^\theta) \\ x_a \sin(\mu_{ab}^\theta) + y_a \cos(\mu_{ab}^\theta) \end{matrix} \right\rangle .$$

The consistency constraints are then restated as follows (and referred to as *relative constraints* from here on):

- ⋄ $\mu^{\langle x,y \rangle}(a, a) = \langle 0, 0 \rangle$;
- ⋄ $\mu^{\langle x,y \rangle}(a, b) = -\mathcal{T}_{ba}[\mu^{\langle x,y \rangle}(b, a)]$ *(anti-symmetry)*;
- ⋄ $\mu^{\langle x,y \rangle}(a, c) = \mu^{\langle x,y \rangle}(a, b) + \mathcal{T}_{ba}[\mu^{\langle x,y \rangle}(b, c)]$ *(additivity)*.

The following sections describe the learning algorithm and the initialization procedure. For clarity and brevity, proofs and a lot of technical detail are omitted, and we concentrate on enforcing the global constraints rather than the relative ones. The extension is straightforward, and the results reported in Section 6 were indeed obtained under *relative* coordinate systems. The complete proofs, treatment of the relative constraints, extension to complete POMDPs and further results can be found in [Sha99].

## 4  Learning the Model

The learning algorithm starts from an initial model $\lambda_0$ and is given a sequence of *experience* E; it returns a revised model $\lambda$, with the goal of maximizing the likelihood $P(\mathsf{E}|\lambda)$. The experience sequence E is of length $T$; each element, $\mathsf{E}_t$, is a pair $\langle r_t, V_t \rangle$, where $r_t$ is the observed relation vector along the $x$, $y$ and $\theta$ dimensions, between the states $q_{t-1}$ and $q_t$ and $V_t$ is the observation vector at time $t$.

Our algorithm extends the Baum-Welch algorithm [Rab89] to deal with the relational information and the factored observation sets. The Baum-Welch algorithm is an expectation-maximization (EM) algorithm [DLR77]; it alternates between

- the *E-step* of computing the state-occupation and state-transition probabilities, $\gamma$ and $\xi$, at each time in the sequence given E and the current model $\lambda$, and

- the *M-step* of finding a new model, $\bar{\lambda}$, that maximizes $P(\mathsf{E}|\lambda, \gamma, \xi)$,

providing monotone convergence of the likelihood function $P(\mathsf{E}|\lambda)$ to a local maximum.

However, our extension introduces an additional component, namely, the relation ($R$) matrix. It can be viewed as having two kinds of observations: *state* observations (as the ordinary HMM — with the distinction that we observe integer vectors rather than integers) and *transition* observations (the odometry relations between states). The latter must satisfy geometrical constraints. Hence, an extension of the standard update formulae, as described below, is required.

### 4.1 State-Occupation Probabilities

Following Rabiner [Rab89], we first compute the forward ($\alpha$) and backward ($\beta$) matrices. $\alpha_t(i)$ denotes the probability density value of observing $E_0$ through $E_t$ and $q_t = s_i$, given $\lambda$; $\beta_t(i)$ is the probability density of observing $E_{t+1}$ through $E_{T-1}$ given $q_t = s_i$ and $\lambda$.

The forward procedure for calculating the $\alpha$ matrix is initialized with
$$\alpha_0(i) = \begin{cases} b_0^i & \text{if } \pi_i = 1 \\ 0 & \text{otherwise} \end{cases}$$
and continued for $0 < t \leq T-1$ with
$$\alpha_t(j) = \sum_{i=0}^{N-1} \alpha_{t-1}(i) A_{i,j} f(r_t | R_{i,j}) b_t^j \ .$$

$f(r_t | R_{i,j})$ denotes the *density* at point $r_t$ according to the distribution represented by the means and variances in entry $i,j$ of the relation matrix $R$, and $b_t^j$ is the probability of observing vector $v_t$ in state $s_j$; that is, $b_t^j = \prod_{i=0}^{l} B_{i,j,v_t[i]}$.

The backward procedure for calculating the $\beta$ matrix is initialized with $\beta_{T-1}(j) = 1$, and continued for $0 \leq t < T-1$ with
$$\beta_t(i) = \sum_{j=0}^{N-1} \beta_{t+1}(j) A_{i,j} f(r_{t+1} | R_{i,j}) b_{t+1}^j \ .$$

Given $\alpha$ and $\beta$, we now compute the state-occupation and state-transition probabilities, $\gamma$ and $\xi$. The state-occupation probabilities are computed as follows:
$$\gamma_t(i) = \Pr(q_t = s_i | E, \lambda) = \frac{\alpha_t(i)\beta_t(i)}{\sum_{j=0}^{N-1} \alpha_t(j)\beta_t(j)} \ ,$$

Similarly, the state-transition probabilities are computed as:
$$\xi_t(i,j) = \Pr(q_t = s_i, q_{t+1} = s_j | E, \lambda)$$
$$= \frac{\alpha_t(i) A_{i,j} b_{t+1}^j f(r_{t+1}|R_{i,j}) \beta_{t+1}(j)}{\sum_{i=0}^{N-1} \sum_{j=0}^{N-1} \alpha_t(i) A_{i,j} b_{t+1}^j f(r_{t+1}|R_{i,j}) \beta_{t+1}(j)} \ .$$

These are essentially the same formulae appearing in Rabiner's tutorial [Rab89], but they also take into account the density of the odometric relations.

In the next phase of the algorithm, the goal is to find a new model, $\overline{\lambda}$, that maximizes $\Pr(E|\lambda, \gamma, \xi)$. Usually, this is simply done using maximum-likelihood estimation of the probability distributions in $A$ and $B$ by computing expected transition and observation frequencies. In our model we must also compute a new relation matrix, $R$, under the constraint that it remain geometrically consistent. Through the rest of this section we use the notation $\overline{v}$ to denote a reestimated value, where $v$ denotes the current value.

### 4.2 Updating Transition and Observation Parameters

The $A$ and $B$ matrices can be straightforwardly reestimated; $\overline{A}_{i,j}$ is the expected number of transitions from $s_i$ to $s_j$ divided by the expected number of transitions from $s_i$; $\overline{B}_{i,j,o}$ is the expected number of times $o$ is observed along the $i$th dimension when in state $s_j$, divided by the expected number of times of being in $s_j$:

$$\overline{A}_{i,j} = \frac{\sum_{t=0}^{T-2} \xi_t(i,j)}{\sum_{t=0}^{T-2} \gamma_t(i)} \ , \quad \overline{B}_{i,j,o} = \frac{\sum_{t=0}^{T-1} I[V_t[i]=o] \gamma_t(j)}{\sum_{t=0}^{T-1} \gamma_t(i)} \ ,$$

where $I[c]$ is an indicator function with value 1 if $c$ is true and 0 otherwise.

### 4.3 Updating Relations Parameters

When reestimating the relation matrix, $R$, the geometrical constraints induce interdependencies among the optimal mean estimates as well as between optimal variance estimates and mean estimates. Parameter estimation under this form of constraints is almost untreated in main-stream statistics [Bar84] and we found no previous existing solutions to the estimation problem faced here. As an illustration, consider the following constrained estimation problem of 2 normal means.

**Example 1** *Consider two sample sets of points $P = \{p_1, p_2, \ldots, p_n\}$ and $Q = \{q_1, q_2, \ldots, q_k\}$, independently drawn from two distinct normal distributions with means $\mu_P$, $\mu_Q$ and variances $\sigma_P^2$, $\sigma_Q^2$, respectively. We are asked to find maximum likelihood estimates for the two distribution parameters. Moreover, we are told that the means of the two distributions are related, such that $\mu_Q = -\mu_P$, as illustrated in Figure 4. If not for the latter constraint, the task is simple [DeG86], and we have:*

$$\mu_P = \frac{\sum_{i=1}^{n} p_i}{n} \ , \quad \sigma_P^2 = \frac{\sum_{i=1}^{n}(p_i - \mu_x)^2}{n} \ ,$$

*and similarly for $\mu_Q$ and $\sigma_Q^2$. However, the constraint $\mu_P = -\mu_Q$ requires finding a single mean $\mu$ and setting the other one to its negated value, $-\mu$. Intuitively, when choosing such a maximum likelihood single mean, the more concentrated sample should have more effect while the more varied sample should be more "submissive". Thus, the overall sample deviation from the means would be minimized and the likelihood of the data – maximized. Therefore, there exists* mutual dependence *between the estimation of the mean and the estimation of the variance.*

*Since the samples are independently drawn, by taking the derivatives of their joint log-likelihood function, with respect to $\mu_P$, $\sigma_P$ and $\sigma_Q$, and equating them to 0, while using the constraint $\mu_Q = -\mu_P$, we obtain the following set of mutual equations for maximum likelihood estimators:*

$$\mu_P = \frac{(\sigma_Q^2 \sum_{i=1}^{n} p_i) - (\sigma_P^2 \sum_{j=1}^{k} q_j)}{n\sigma_Q^2 + k\sigma_P^2} \ , \quad \mu_Q = -\mu_P \ ,$$

$$\sigma_P^2 = \frac{\sum_{i=1}^{n}(p_i - \mu_P)^2}{n} \ , \quad \sigma_Q^2 = \frac{\sum_{j=1}^{k}(q_j + \mu_P)^2}{k} \ .$$

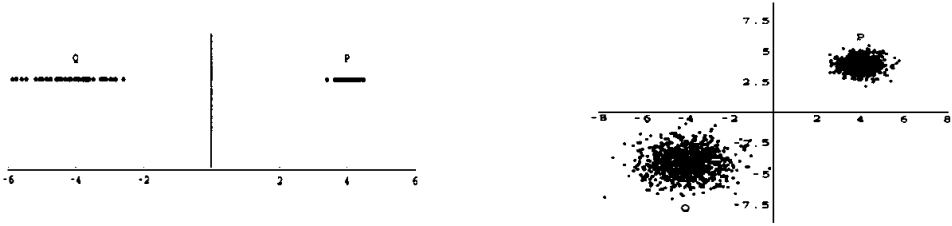

Figure 4: Examples of two sets of normally distributed points with constrained means, in 1 and in 2 dimensions.

*By substituting the expressions for $\sigma_P$ and $\sigma_Q$ into the expression for $\mu_P$, we obtain a cubic equation which is cumbersome, but still solvable (in this simple case). The solution provides a maximum likelihood estimate for the mean and variance under the constraint $\mu_Q = -\mu_P$.* □

We now proceed to the actual update of the relation matrix under constraints. For clarity, we initially discuss only the first two geometrical constraints, and discuss the additivity constraint in Section 4.4. Recall that we concentrate here on the global constraints enforcement, although the same idea is applied for the state-relative constraints.

*Zero distances* between states and themselves are trivially enforced, by setting all the diagonal entries in the $R$ matrix to 0, with a small variance.

*Anti-symmetry* within a global coordinate system is enforced by using the data recorded along the transition from $s_j$ to $s_i$ as well as from $s_i$ to $s_j$ when reestimating $\mu(R_{i,j})$. As shown in Example 1, the variance has to be taken into account, leading to the following set of mutual equations:

$$\overline{\mu}_{i,j}^m = \frac{\sum_{t=0}^{T-2}\left[\frac{r_t[m]\xi_t(i,j)}{(\overline{\sigma}_{i,j}^m)^2} - \frac{r_t[m]\xi_t(j,i)}{(\overline{\sigma}_{j,i}^m)^2}\right]}{\sum_{t=0}^{T-2}\left[\frac{\xi_t(i,j)}{(\overline{\sigma}_{i,j}^m)^2} + \frac{\xi_t(j,i)}{(\overline{\sigma}_{j,i}^m)^2}\right]}, \quad (1)$$

$$(\overline{\sigma}_{i,j}^m)^2 = \frac{\sum_{t=0}^{T-2}[\xi_t(i,j)(r_t[m] - \overline{\mu}_{i,j}^m)^2]}{\sum_{t=0}^{T-2}\xi_t(i,j)}. \quad (2)$$

For the $x$ and $y$ dimensions this amounts to a complicated but still solvable cubic equation. However, in the more general case, when accounting for the orientation of the robot, and also when complete additivity is enforced, we do not obtain such closed form reestimation formulae.

To avoid these reestimation hardships, we use a *lag-behind* update rule; the yet-unupdated estimate of the variance is used for calculating a new estimate for the mean, and this new mean estimate is used to update the variance, using Eq. 2. [2] Thus, the mean is updated using a variance parameter that *lags behind* it in the update process, and the reestimation formula 1 needs to use $\sigma_{i,j}^m$ rather than $\overline{\sigma}_{i,j}^m$:

$$\overline{\mu}_{i,j}^m = \frac{\sum_{t=0}^{T-2}\left[\frac{r_t[m]\xi_t(i,j)}{(\sigma_{i,j}^m)^2} - \frac{r_t[m]\xi_t(j,i)}{(\sigma_{j,i}^m)^2}\right]}{\sum_{t=0}^{T-2}\left[\frac{\xi_t(i,j)}{(\sigma_{i,j}^m)^2} + \frac{\xi_t(j,i)}{(\sigma_{j,i}^m)^2}\right]}. \quad (3)$$

This lag-behind policy is an instance of generalized EM [Sha99], which guarantees monotone convergence to a local maximum of the likelihood function.

[2] A similar approach, termed *one step late* update, is taken by others applying EM to highly non-linear optimization problems [MK97].

Similarly, the reestimation formula for the von Mises mean and concentration parameters of the *heading change* between states $s_i$ and $s_j$ is the solution to the equations:

$$\overline{\mu}_{i,j}^\theta = \arctan\left(\frac{\sum_{t=0}^{T-2}[\sin(r_t[\theta])(\xi_t(i,j)\overline{\kappa}_{i,j} - \xi_t(j,i)\overline{\kappa}_{j,i})]}{\sum_{t=0}^{T-2}[\cos(r_t[\theta])(\xi_t(i,j)\overline{\kappa}_{i,j} + \xi_t(j,i)\overline{\kappa}_{j,i})]}\right)$$

$$\frac{I_1[\overline{\kappa}_{i,j}^\theta]}{I_0[\overline{\kappa}_{i,j}^\theta]} = \max\left[\frac{\sum_{t=0}^{T-2}[\xi_t(i,j)\cos(r_t[\theta] - \overline{\mu}_{i,j}^\theta)]}{\sum_{t=0}^{T-2}\xi_t(i,j)}, 0\right]. \quad (4)$$

Again, to avoid the need to solve these mutual equations, we take advantage of the "lag-behind" strategy, updating the mean using the current estimates of the concentration parameters, $\kappa_{i,j}$, $\kappa_{j,i}$, as follows:

$$\overline{\mu}_{i,j}^\theta = \arctan\left(\frac{\sum_{t=0}^{T-2}[\sin(r_t[\theta])(\xi_t(i,j)\kappa_{i,j} - \xi_t(j,i)\kappa_{j,i})]}{\sum_{t=0}^{T-2}[\cos(r_t[\theta])(\xi_t(i,j)\kappa_{i,j} + \xi_t(j,i)\kappa_{j,i})]}\right),$$

and then calculating the new concentration parameters based on the newly updated mean, as the solution to equation 4, through the use of lookup-tables.

A possible alternative to our lag-behind approach is to update the mean as though the assumption $\sigma_{j,i} = \sigma_{i,j}$ holds. Under this assumption, the variance terms in equation 1 cancel out, and the mean update is independent of the variance once again. Then the variances are updated as stated in equation 2, *without* assuming any constraints over them. This approach was taken in earlier stages of this work [SK97, SK98]. The lag-behind strategy is superior, both according to our experiments, and due to its being an instance of generalized EM.

### 4.4 Enforcing Additivity

Note that the additivity constraint implies the other two geometrical constraints, thus enforcing it results in complete geometrical consistency. One way to enforce additivity is by using the iterative anti-symmetric update procedure described above, augmenting each iteration with a procedure for deriving an additive model from the anti-symmetric one. Our experience with such a technique proved unsatisfactory, typically converging to poor models or altogether failing to converge.

We briefly describe here the method for directly enforcing additivity through the reestimation procedure. As before, we restrict the discussion to global coordinate systems.

### 4.4.1 Additivity in the $x$, $y$ dimensions

The main observation underlying our approach is that the additivity constraint is a result of the fact that states can be embedded in a *geometrical space*. That is, assuming we have $N$ states, $s_0, \ldots, s_{N-1}$, there are points on the $X$, $Y$ and $\theta$ axes, $x_0, \ldots, x_{N-1}, y_0, \ldots, y_{N-1}, \theta_0, \ldots, \theta_{N-1}$, respectively, such that each state, $s_i$, is associated with the coordinates $\langle x_i, y_i, \theta_i \rangle$. Assuming one global coordinate system, the mean odometric relation from state $s_i$ to state $s_j$ can be expressed as: $\langle x_j - x_i, y_j - y_i, \theta_j - \theta_i \rangle$.

During the *maximization* phase of the EM iteration, rather than try to maximize with respect to $N^2$ *odometric relation vectors*, $\langle \mu_{ij}^X, \mu_{ij}^Y, \mu_{ij}^\theta \rangle$, we *reparameterize* the problem. Specifically, we express each odometric relation as a function of two of the $N$ *state positions*, and maximize with respect to the *unconstrained, $N$ state positions*. For instance, for the $X$ dimension, we find during the maximization step $N$ 1-dimensional points, $x_0, \ldots, x_{N-1}$, from which we calculate $\mu_{ij}^x = x_j - x_i$. Moreover, since all we are interested in is finding the best *relationships* between $x_i$ and $x_j$, we can fix one of the $x_i$'s at 0 (e.g. $x_0 = 0$), and only find optimal estimates for the other $N-1$ state positions. The variance reestimation remains as before, and the lag-behind policy is used to eliminate the interdependency between the update of the mean and the variance parameters.

### 4.4.2 Additive Heading Estimation

Unfortunately, the reparameterization described above is not feasible for heading change estimation, due to the von Mises distribution assumption over the heading measures; By reparameterizing $\mu_{ij}^\theta$ as $\theta_j - \theta_i$ and maximizing the likelihood function with respect to the $\theta$'s, we obtain a set of $N-1$ *trigonometric equations* with terms of the form $\cos(\theta_j) \cdot \sin(\theta_i)$ which do not enable simple solution.

A possible alternative is to use the anti-symmetric reestimation procedure, followed by a perpendicular *projection* operator, mapping the headings vector $\langle \mu_{00}^\theta, \ldots, \mu_{ij}^\theta, \ldots, \mu_{N-1,N-1}^\theta \rangle$, $0 \le i, j \le N-1$, which does not satisfy additivity, onto a vector of headings within an *additive linear vector space*. Simple orthogonal projection is not satisfactory within our setting, since it simply looks for the additive vector closest to the non-additive one, ignoring the fact that some of the entries in the non-additive vector are based on a lot of observations, while others are based on hardly any data at all. Intuitively, we would like to keep the estimates that are well accounted for intact, and adapt the less accounted for estimates in order to meet the additivity constraint. More precisely, we would like to project the *non-additive* heading estimates vector onto a *subspace* of the *additive* vector space, in which the vectors have the same values as the non-additive vector in the entries that are well-accounted for. The culprit is that the latter subspace is *not* a *linear* vector space (for instance, it does not satisfy closure under scalar multiplication), and the projection operator over linear spaces can not be ap-plied directly. Still, this set of vectors does form an *affine* vector space, and we can project onto it using a special technique from linear algebra, as explained below [3].

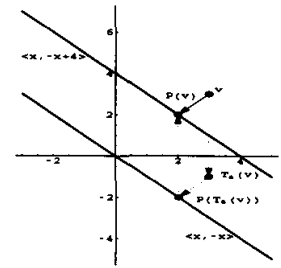

**Figure 5**: Projecting $v$ onto an affine space

**Definition**    $A \subseteq \mathcal{R}^n$ *is an n-dimensional affine space if for all vectors* $v_a \in A$, *the set of vectors:* $A - v_a \stackrel{def}{=} \{u_a - v_a | u_a \in A\}$ *is a linear space.*

Hence, we can pick a vector, $v_{a_1} \in A$, and define the translation $T_a : A \to V$, where $V$ is a linear space, $V = A - v_{a_1}$. This translation is trivially extended for any vector $v' \in \mathcal{R}^n$, by defining $T_a(v') = v' - v_{a_1}$. In order to project a vector $v \in R^n$ onto A, we apply the translation $T_a$ to $v$ and project $T_a(v)$ onto $V$, which results in a vector $\mathcal{P}(T_a(v))$ in $V$. By applying the inverse transform $T_a^{-1}$ to it, we obtain the projection of $v$ on A, as demonstrated in Figure 5. The linear space in the figure is the two dimensional vector space $\{\langle x, y \rangle | y = -x\}$, and the affine space is $\{\langle x, y \rangle | y = -x + 4\}$. The transform $T_a$ consists of subtracting the vector $\langle 0, 4 \rangle$. The solid arrow corresponds to the direct projection of the vector $v$ onto the point $\mathcal{P}(v)$ of the affine space. The dotted arrows represent the projection via translation of $v$ to $T_a(v)$, the projection of the latter onto the linear vector space, and the inverse translation of the result, $\mathcal{P}(T_a(v))$, onto the affine space.

Although the procedure for preserving additivity over headings is not proven to preserve monotone convergence of the likelihood function towards a local maximum, our extensive experiments consisting of hundreds of runs have shown that monotone convergence is preserved.

## 5 Choosing an Initial Model

Typically, in instances of the Baum-Welch algorithm, an initial model is picked uniformly at random from the space of all possible models, perhaps trying multiple initial models to find different local likelihood maxima. An alternative approach we have reported [SK97] was based on clustering the accumulated odometric information using the simple k-means algorithm [DH73], taking the clusters to be the states in which the observations were recorded, to obtain state and observation counts and estimate the model parameters.

When perpendicularity is assumed, as shown in Figure 1, the k-means algorithm assigns the same cluster (state) to odometric readings recorded at close locations, leading to reasonable initial models. However, when this assump-tion is dropped, as illustrated in Figure 2, the cumulative rotational error distorts the odometric location recorded within a global coordinate system, so that the location assigned to the same state during multiple visits varies greatly and would not be recognized as "the same" by a simple location-based clustering algorithm. To overcome this, we developed an alternative initialization heuristics, based *directly* on the recorded *relations* between states – rather than on absolute states location. For clarity, the description here is informal, consisting mostly of an illustrative example, and enforcing global consistency constraints.

---
[3]Many thanks to John Hughes for introducing this technique.

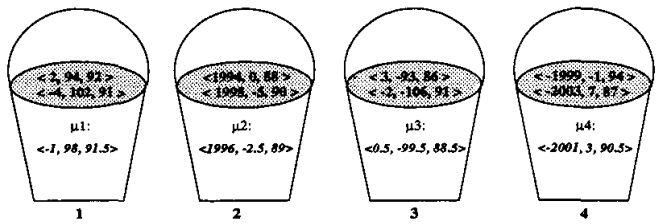

**Figure 6**: The bucket assignment of the example sequence.

Given a sequence of observations and odometric readings E, we begin by clustering the odometric readings into *buckets*. The number of buckets is at most the number of distinct state transitions recorded in the sequence. The goal at this stage is to have each bucket contain all the odometric readings that are close to each other along all three dimensions.

To achieve this, we start by fixing a predetermined, small standard deviation value along the $x$, $y$, and $\theta$ dimensions. Denote these standard deviation values $\sigma_x, \sigma_y, \sigma_\theta$ respectively, (typically $\sigma_x = \sigma_y$). The first odometric reading is assigned to bucket 0 and the mean of this bucket is set to be the value of this reading. Through the rest of the process the subsequent odometric readings are examined. If the next reading is within 1.5 standard deviations along each of the three dimensions from the mean of some existing non-empty bucket, add it to the bucket and update the bucket mean accordingly. If not, assign it to an empty bucket and set the mean of the bucket to be this reading.

This algorithm guarantees that all the odometric readings in each bucket are within a range of $1.5 \cdot \langle \sigma_x, \sigma_y, \sigma_\theta \rangle$ from the bucket mean. Since the actual *sample* standard deviation of each bucket can not exceed the fixed deviation used during the bucketing process, intuitively, each bucket is tightly concentrated about its mean. We note that other clustering algorithms [DH73] could be used at the bucketing stage.

**Example 2** *We would like to learn a 4-state model from a sequence whose odometric component is as follows:*

$\langle 2\ 94\ 92\rangle, \langle 1994\ 0\ 88\rangle, \langle 3\ -93\ 86\rangle, \langle -1999\ 1\ 94\rangle,$
$\langle -4\ 102\ 91\rangle, \langle 1998\ -5\ 90\rangle, \langle -2\ -106\ 91\rangle, \langle -2003\ 7\ 87\rangle.$

*As a first stage we place these readings into buckets. Suppose the standard deviation constant is 20. The placement is as shown in Figure 6. The mean value associated with each bucket is shown as well.*    □

The next stage of the algorithm is the *state-tagging* phase, in which each odometric reading, $r_t$, is assigned a pair of states, $s_i$, $s_j$, denoting the origin state (from which the transition took place) and the destination state (to which the transition led), respectively. In conjunction, the mean entries, $\mu_{ij}$, of the relation matrix, $R$, are populated.

**Example 2 (cont.)** *Returning to the sequence above, the process is demonstrated in Figure 7. We assume that the data recording starts at state 0, and that the odometric change through self transitions is 0, with some small standard deviation (we use 20 here as well). This is shown on part A of the figure.*

*Since the first element in the sequence, $\langle 2\ 94\ 92\rangle$, is more than two standard deviations away from the mean $\mu[0][0]$ and no other entry in the relation row of state 0 is populated, we pick 1 as the next state and populate the mean $\mu[0][1]$ to be the same as the mean of bucket 1, to which $\langle 2\ 94\ 92\rangle$ belongs. To maintain geometrical consistency the mean $\mu[1][0]$ is set to be $-\mu[0][1]$, as shown in part B of the figure. We now have populated 2 off-diagonal entries, and the state sequence is $\langle 0, 1 \rangle$. The entry $[0][1]$ in the matrix becomes associated with bucket 1, and this information is recorded for helping with tagging future odometric readings belonging to the same bucket.*

*The next odometric reading, $\langle 1994\ 0\ 88\rangle$, is a few standard deviations from any populated mean in row 1 (where 1 is the current believed state). Hence, we pick a new state 2, and set the mean $\mu[1][2]$ to be $\mu 2$ — the mean of bucket 2 — to which the reading belongs (Figure 7 C). The entry $[1][2]$ is recorded as associated with bucket 2. To preserve antisymmetry and additivity, $\mu[2][1]$ is set to $-\mu[1][2]$. $\mu[0][2]$ is set to be the sum $\mu[0][1] + \mu[1][2]$, and $\mu[2][0]$ is set to $-\mu[0][2]$. Similarly, $\mu[2][3]$ is updated to be the mean of bucket 3, causing the setting of $\mu[3][2]$, $\mu[1][3]$, $\mu[0][3]$, $\mu[3][1]$, and $\mu[3][0]$. Bucket 3 is associated with $\mu[2][3]$.*

*At this stage the odometric table is fully populated, as shown in part D of Figure 7. The state sequence at this point is: $\langle 0, 1, 2, 3 \rangle$. The next reading, $\langle -1999\ -1\ 94\rangle$, is within one standard deviation from $\mu[3][0]$ and therefore the next state is 0. Entry $[3][0]$ is associated with bucket 4, (the bucket to which the reading was assigned), and the state sequence becomes: $\langle 0, 1, 2, 3, 0\rangle$.*

*The next reading, being from bucket 1, is associated with the relation from state 0 that is tagged by bucket 1, namely, state 1. By repeating this for the last two readings, the final state transition sequence becomes $\langle 0, 1, 2, 3, 0, 1, 2, 3, 0\rangle$.*    □

Once the state-transition sequence is obtained, the rest of the initialization algorithm is the same as it is for k-means based initialization, deriving state-transition counts from the state-transition sequence, assigning the observations to the states under the assumption that the state sequence is correct, and obtaining state-transition and observation probabilities. The initialization phase does not incur much computational overhead, and is equivalent time-wise to performing one additional iteration of the EM procedure.

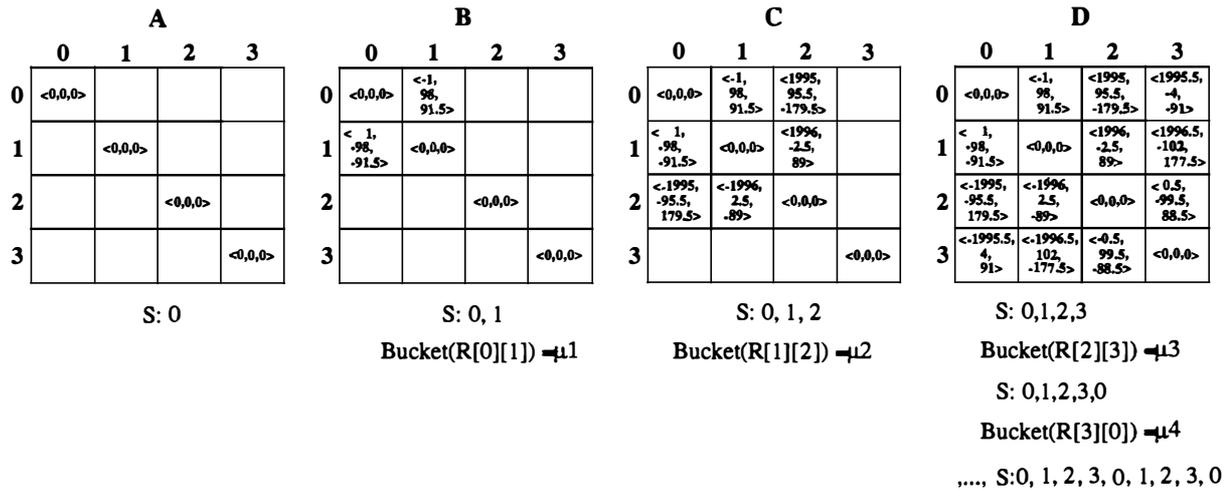

Figure 7: Populating the odometric relation matrix and creating a state tagging sequence.

## 6 Experiments and Results

Our experiments consist of learning models from both real and simulated robot data (*without* assuming perpendicularity), evaluating the results both visually and statistically.

### 6.1 Experimental Setting

We ran our robot, Ramona, along a *prescribed*[4] directed path in the Brown CS department. Low-level routines let Ramona move forward from one hallway intersection to the next and to turn 90° to the left or right. Ultrasonic data interpretation lets her perceive, in three directions – front, left and right – whether there is an open space, a door, a wall, or something unknown. Doors and intersections constitute *states*. When they are detected, Ramona stops and records its observations, and its odometric change between the previous and the current state. All recorded measures as well as the actions are, of course, subject to error.

The path Ramona followed consists of 4 connected corridors, including 17 states, and is shown as an HMM in Figure 8. Black dots represent the physical locations of states. Multiple states (shown as numbers in the plot) associated with a single location correspond to *different orientations* of the robot at that location. The larger circle, at the bottom left corner, represents the initial position. Solid arrows represent the most likely directed transition (corridor traversed) between states and dashed arrows represent transitions that have probability 0.2 or higher (if such exist). The arrow length represents the corridor length, *drawn to scale*. The observations associated with each state are omitted for clarity. A projection of the odometric readings recorded along the $x$ and $y$ dimensions, was shown in Figure 2.

To statistically evaluate our algorithm, we use a simulated office environment in which the robot follows a prescribed path. It is represented as an HMM consisting of 44 states, and the associated transition, observation, and odometric distributions. Figure 11 depicts this HMM. We generated 5 data sequences from the model, each of length 800, using Monte Carlo sampling. One of these sequences is depicted in Figure 12. Again, observations are omitted, and this is a projection of the odometry readings onto a global 2-dimensional coordinate system. For each sequence we ran our algorithm 10 times. For comparison, we also ran the standard Baum-Welch algorithm, not using odometric information, 10 times on each sequence.

### 6.2 Results

We used our algorithm, enforcing additivity and using the initialization procedure of Section 5, to learn a model of the environment from the data gathered by Ramona. Figure 9 depicts a typical model learned from that data; the learned $R$ matrix was used for determining relative state positions. It is clear that the model corresponds well both topologically and geometrically to the true environment. The observation distributions learned are omitted, but they too reflect well the walls, doors and openings encountered, while incorporating the identification error resulting from noisy sensors. Note that the initial state, 0, is not well positioned geometrically with respect to the rest of the model; due to the large number of states neighboring the initial state, 0, in the true environment, it was not recognized that we ever returned to this particular state during the loop. Therefore, only one expected transition was recognized from state 0 to state 1 by the algorithm. When projecting the angles to maintain additivity, the angle from state 0 to 1 was consequently compromised, maintaining the rectangular geometry among the more regularly visited states.

Note that learning such circular topologies is very challenging, since their highly symmetric nature makes it difficult to *distinguish separate* states, as well as to *identify* when the *same* state is revisited; as far as we know no other topological approach can learn such models from raw data, and the only other work which handles them is the grid-based geometrical approach of Thrun et al [TBF98].

Figure 10 shows the topology of a typical HMM learned using the standard Baum-Welch algorithm *without* odometric information. The bold circle represents the initial state. The arrows semantics is as before. The loop topology of the traversed environment is obviously *not* captured.

---
[4]Hence, no decisions are executed by the robot, and the model is an HMM and not a complete POMDP.

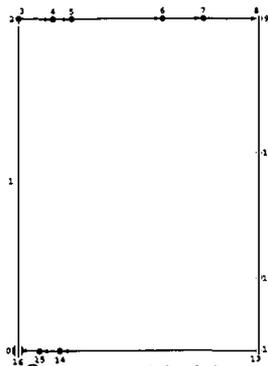

Figure 8: True model of the corridors Ramona traversed.

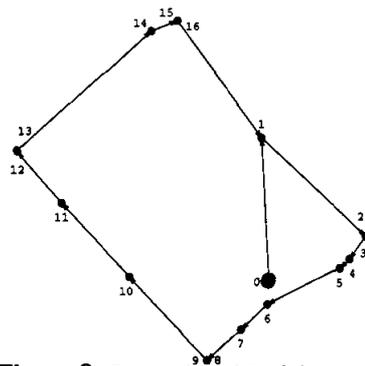

Figure 9: Learned model of the corridors Ramona traversed.

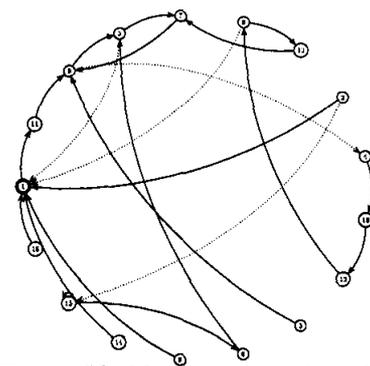

Figure 10: Model learned without the use of odometric information.

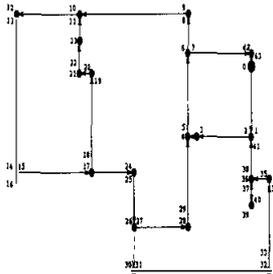

Figure 11: True model of the simulated hallway environment.

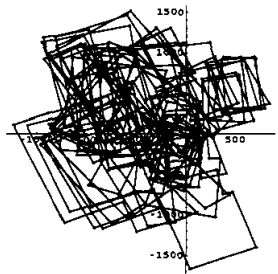

Figure 12: A data sequence generated from the simulated model

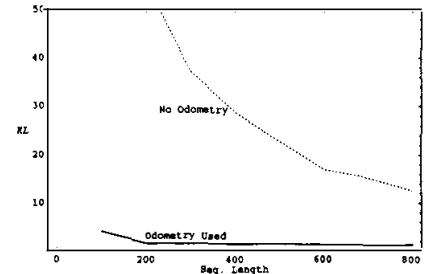

Figure 13: Average KL divergence as a function of length.

Traditionally, in simulation experiments, learned models are quantitatively compared to the actual model that generated the data. Each of the models induces a probability distribution on strings of observations; the Kullback-Leibler divergence [KL51] between the two distributions is a measure of how far the learned model is from the true model. We report our simulation results in terms of a sampled version of the KL divergence, as described by Juang and Rabiner [JR85]. It is based on generating sequences of sufficient length according to the distribution induced by the true model, and comparing their likelihoods according to the learned model, with the true model likelihoods. Odometric information is ignored when applying the KL measure, thus allowing comparison between purely topological models that are learned with and without odometry.

Table 1 lists the KL divergence between the true and learned model, as well as the number of iterations until convergence was reached, for each of the 5 simulation sequences under the two learning settings, averaged over 10 runs per sequence. The table demonstrates that the KL divergence with respect to the true model for models learned using odometric data, is about *8 times smaller* than for models learned without it. To check the significance of our results we used the simple two-sample t-test. The models learned using odometric information have highly statistically significantly ($p \ll 0.0005$) lower average KL divergence than the others. In addition, the number of iterations required for convergence when learning using odometric information is smaller than required when ignoring such information. Again, the t-test verifies the significance ($p < 0.005$) of this result.

Learning HMMs obviously requires visiting states and transitioning between them multiple times, to gather sufficient data for robust statistical modeling. Intuitively, exploiting odometric data can help reduce the number of visits needed for obtaining a reliable model. To examine the influence of reduction in the length of data sequences on the quality of the learned models, we took one of the 5 sequences and used its prefixes of length 100 to 800 (the complete sequence), in increments of 100, as training sequences. We ran the two algorithmic settings over each of the 8 prefix sequences, 10 times repeatedly. The KL divergence was then used to evaluate each resulting model with respect to the true model. For each prefix length we averaged the KL divergence over the 10 runs. Figure 13 depicts the average KL divergence as a function of the sequence length for each of the settings. It demonstrates that, in terms of the KL divergence, our algorithm, using odometric information, is robust in the face of data reduction, (down to 200 data points). In contrast, learning without the use of odometry quickly deteriorates as the amount of data is reduced.

## 7  Conclusions

Odometric information, which is often readily available in the robotics domain, makes it possible to learn hidden Markov models efficiently and effectively, while using shorter training sequences. The odometric information can be directly incorporated into the traditional HMM model, maintaining convergence of the reestimation algorithm to a local maximum of the likelihood function.

Even though we are primarily interested in the underlying topological model (transition and observation probabilities), our experiments demonstrate that using odometric relations can both reduce the number of iterations required by the algorithm and improve the resulting model.

Learning HMMs with Geometrical Constraints    611| Seq. # | | 1 | 2 | 3 | 4 | 5 |
|---|---|---|---|---|---|---|
| With Odo | KL | 1.46 | 1.18 | 1.20 | 1.02 | 1.22 |
| | Iter # | 11.8 | 36.8 | 30.7 | 24.6 | 33.3 |
| No Odo | KL | 6.91 | 9.93 | 10.03 | 9.54 | 12.43 |
| | Iter # | 113.3 | 113.1 | 102.0 | 104.2 | 112.5 |

**Table 1**: Average results of 2 learning settings with 5 training sequences.

The initialization procedure and the enforcement of the additivity constraint over relatively small models prove helpful both topologically and geometrically. An extensive study [Sha99] shows that for long data sequences, generated from large models, enforcing only *anti-symmetry* rather than *additivity*, leads to better topological models. This is because in these cases, initialization is not always good, and additivity may over-constrain the learning to an unfavorable area. Learning of large models may benefit from enforcing only anti-symmetry during the first few iterations, and complete additivity in later iterations. Alternatively, we may use our algorithm to learn separate models for small portions of the environment, combining them later into one complete model.

The work presented here demonstrates how domain-specific information and constraints can be incorporated into the statistical estimation process, resulting in better models, while requiring shorter data sequences. We strongly believe that this idea can be applied in domains other than robotics. In particular, the acquisition of HMMs for use in Molecular Biology may greatly benefit from exploiting geometrical (and other) constraints on molecular structures. Similarly, temporal constraints may be exploited in domains in which POMDPs are appropriate for decision-support, such as air-traffic control and medicine.

## Acknowledgments

I am grateful to Leslie Kaelbling for her guidance throughout this work, and particularly for suggesting the bucketing phase, Sebastian Thrun for his insightful comments, John Hughes for the affine projection technique, Jim Kurien for the low level robot software and Bill Smart for maintaining Ramona.## References

[Asa91]  M. Asada, Map Building for a Mobile Robot from Sensory Data, *Autonomous Mobile Robots*, S. S. Iyengar and A. Elfes, eds., pp. 312–322, IEEE Press, 1991.

[Bar84]  R. Bartels, Estimation in a Bidirectional Mixture of von Mises Distributions, *Biometrics*, 40, pp. 777–784, 1984.

[CKK96]  A. R. Cassandra, L. P. Kaelbling and J. A. Kurien, Acting Under Uncertainty: Discrete Bayesian Models for Mobile-Robot Navigation, *Proc. of the Int. Conf. on Intelligent Robots and Systems*, 1996.

[DeG86]  M. H. DeGroot, *Probability and Statistics*, Addison-Wesley, 2nd edn., 1986.

[DH73]  R. O. Duda and P. E. Hart, *Pattern Classification and Scene Analysis*, chap. 6, John Wiley and Sons, 1973.

[DLR77]  A. P. Dempster, N. M. Laird and D. B. Rubin, Maximum Likelihood from Incomplete Data via the EM Algorithm, *Journal of the Royal Statistical Society*, 39 (1), pp. 1–38, 1977.

[GGD53]  E. G. Gumbel, J. A. Greenwood and D. Durand, The Circular Normal Distribution: Theory and Tables, *American Statistical Society Journal*, 48, pp. 131–152, 1953.

[GJ97]  Z. Ghahramani and M. I. Jordan, Factorial Hidden Markov Models, Proc. of the *Int. Conf. on Machine Learning*, 1997.

[HG95]  D. Heckerman and D. Geiger, Learning Bayesian Networks: A Unification for Discrete and Gaussian Domains, *Proc. of the Int. Conf. on Uncertainty in AI*, pp. 274–284, 1995.

[JR85]  B. H. Juang and L. R. Rabiner, A Probabilistic Distance Measure for Hidden Markov Models, *AT&T Technical Journal*, 64 (2), pp. 391–408, 1985.

[KB91]  B. Kuipers and Y.-T. Byun, A Robot Exploration and Mapping Strategy Based on a Semantic Hierarchy of Spatial Representations, *Journal of Robotics and Autonomous Systems*, 8, pp. 47–63, 1991.

[KL51]  S. Kullback and R. A. Leibler, On Information and Sufficiency, *Annals of Mathematical Statistics*, 22 (1), pp. 79–86, 1951.

[KS96a]  S. Koenig and R. G. Simmons, Passive Distance Learning for Robot Navigation, *Proc. of the Int. Conf. on Machine Learning*, pp. 266–274, 1996.

[KS96b]  S. Koenig and R. G. Simmons, Unsupervised Learning of Probabilistic Models for Robot Navigation, *Proc. of the Int. Conf. on Robotics and Automation*, 1996.

[Mar72]  K. V. Mardia, *Statistics of Directional Data*, Academic Press, 1972.

[ME85]  H. P. Moravec and A. Elfes, High Resolution Maps from Wide Angle Sonar, *Proc. of the Int. Conf. on Robotics and Automation*, pp. 116–121, 1985.

[MK97]  G. J. McLachlan and T. Krishnan, *The EM Algorithm and Extensions*, John Wiley & Sons, 1997.

[NPB95]  I. Nourbakhsh, R. Powers and S. Birchfield, DERVISH: An Office-Navigating Robot, *AI Magazine*, 16 (1), pp. 53–60, 1995.

[Rab89]  L. R. Rabiner, A Tutorial on Hidden Markov Models and Selected Applications in Speech Recognition, *Proc. of the IEEE*, 77 (2), pp. 257–285, 1989.

[Sha99]  H. Shatkay, *Learning Models for Robot Navigation*, Ph.D. thesis, Tech. Rep. CS-98-11, Dept. of Computer Science, Brown University, Providence, RI, 1999.

[SK95]  R. G. Simmons and S. Koenig, Probabilistic Navigation in Partially Observable Environments, in *Proc. of the Int. Joint Conf. on Artificial Intelligence*, 1995.

[SK97]  H. Shatkay and L. P. Kaelbling, Learning Topological Maps with Weak Local Odometric Information, in *Proc. of the Int. Joint Conf. on Artificial Intelligence*, 1997.

[SK98]  H. Shatkay and L. P. Kaelbling, Heading in the Right Direction, in *Proc. of the Int. Conf. on Machine Learning*, 1998.

[SSC91]  R. Smith, M. Self and P. Cheeseman, A Stochstic Map for Uncertain Spatial Relationships, in *Autonomous Mobile Robots*, S. S. Iyengar and A. Elfes, eds., pp. 323–330, IEEE Computer Society Press, 1991.

[TBF98]  S. Thrun, W. Burgard and D. Fox, A Probabilistic Approach to Concurrent Map Acquisition and Localization for Mobile Robots, *Machine Learning*, 31, pp. 29–53, 1998.